\documentclass[letterpaper]{article}
\usepackage{calc,amsmath,amssymb,amsfonts}
\usepackage[LGR,T1]{fontenc}
\usepackage[greek,english]{babel}
\usepackage{xcolor,longfbox,fancyhdr}
\usepackage[margin=1in,noheadfoot]{geometry}
\usepackage{enumitem,hyperref}
\hypersetup{colorlinks=true,allcolors=blue}
\usepackage[pdftex]{graphicx}
\makeatletter\newdimen\@tempdimd\makeatother
\fancypagestyle{Standard}{\fancyhf{}
  \fancyhead[L]{}
  \fancyfoot[L]{}

}
\begin{document}
\clearpage
\pagestyle{Standard}
\setcounter{page}{1}

\section{\textbf{Detecting Emotion Drift in Mental Health Text Using Pre-Trained Transformers}}

\bigskip
{\color[HTML]{434343}

\subsection{\textbf{\textcolor{black}{Author}}}}

Shibani Sankpal 

\subsection{\textbf{\textcolor{black}{Abstract}}}

This study investigates emotion drift: the change in emotional state across a single text, within mental health-related
messages. While sentiment analysis typically classifies an entire message as positive, negative, or neutral, the
nuanced shift of emotions over the course of a message is often overlooked. This study detects sentence-level emotions
and measures emotion drift scores using pre-trained transformer models such as DistilBERT and RoBERTa. The results provide insights into patterns of emotional escalation or relief in mental health
conversations. This methodology can be applied to better understand emotional dynamics in content.

\bigskip

{\color[HTML]{434343}
\subsection{\textbf{\textcolor{black}{1. Introduction}}}}

\bigskip

Understanding human emotions in written communication has become increasingly important, especially in the context of
mental health. More traditional methods to\enskip{}understand sentiment classify whole texts for being positive,
negative, or neutral and provide a high level overview of the emotional tone (Dilmegani, 2024). However, these
algorithms do not consider the change or drift in emotional state within a single text. These types of large
emotion\enskip{}shifts, or drifts in emotions can be indicative of emotion escalation or instability, which is
particularly relevant in mental health(Mitchell, 2021).

\bigskip

Recent improvements in natural language\enskip{}processing such as pre-trained transformer models (e.g., DistilBERT and
RoBERTa) make it possible to analyze emotions which are more subtle on the sentence level(Acheampong et al., 2020). By using these models,
the\enskip{}ability to examine subtle emotional transitions and measure changes of emotions in texts is made possible.
Comprehending these dynamics can increase early detection of emotional distress, improve automated mental health care
systems, and offer deeper insights into users' emotional experiences.

\bigskip

In this study, I propose a framework for detecting emotion drift in mental health texts. I have analyzed sentence-level
emotions using transformer models, compute emotion drift scores, and evaluate their ability to capture emotional
variability. Our results demonstrate how emotional analysis can uncover patterns of emotional escalation that are not
captured by traditional sentiment analysis.A lightweight Streamlit application was also developed to demonstrate the
practical use of emotion drift analysis, allowing users to input any text and instantly view sentence-level emotions,
the emotion timeline, drift score, and overall sentiment in an intuitive and interactive interface.

\bigskip
\bigskip

{\color[HTML]{434343}
\subsection{\textbf{\textcolor{black}{2. Literature Review}}}}

\bigskip

Emotion analysis in text has become a relevant area of research from natural language processing (NLP),
particularly\enskip{}within areas related to human welfare, online communication and mental health(Plaza-del Arco et
al., 2024). Most of the existing sentiment analysis\enskip{}models are built in such a way that they assign one global
sentiment (positive, negative or neutral) to an entire message. This has worked well in various scenarios such as
product review, social media polarity classification but does not model the fine-grained emotional dynamics commonly
observed in user-generated content(Pang \&\enskip{}Lee, 2008). Psychological evidence suggests that people often
experience ambivalent or contradictory\enskip{}feelings when listening to a single story, especially when participants
are discussing something negative about their life or psychological problems (Liu, 2024). The recent advancements in
the field, such as transformers and deep\enskip{}learning architectures, have allowed for classification of these
fine-grained emotions. 

\bigskip

The utilization of pre-trained models such as BERT (Devlin\enskip{}et al.,n.d.), RoBERTa and their distilled versions
can improve the recognition performance of different emotion types, including joy, anger, sadness, fear and surprise.
The availability of such rich annotations has also expanded the information available for emotion classification
research, by modularization problems that benefit from richer annotations like those found in GoEmotions(GoEmotions: A
Dataset for Fine-Grained Emotion Classification, n.d.) with\enskip{}27 emotion labels. Other data sets such as the
Emotion dataset \enskip{}used in this work include a selected few core emotions, allowing for simpler benchmarking
while maintaining emotional variety.However, most of these models and datasets still evaluate emotions at the document
level, leaving sentence-level emotional transitions largely unexplored.

\bigskip

The concept of emotion drift, defined as the change in emotional state across different segments of a single piece of
text, has only recently started receiving attention. A few studies have explored emotional trajectories in narratives
and long-form posts, showing that measuring emotional progression can uncover latent psychological indicators and
provide deeper insights into user behaviour(Christ et al., 2024). In mental health contexts, emotional fluctuations
have been linked to stress, anxiety, mood instability, and coping mechanisms, making them valuable for early detection
and support frameworks. Despite this, there is limited work on operationalising emotion drift using pre-trained
transformer models, particularly in real-time, user-facing applications.

\bigskip

To address this gap, recent research efforts have focused on applying transformer-based emotion classifiers at the
sentence level to compute emotional trajectories and drift scores. Models such as DistilBERT and DistilRoBERTa provide
a computationally efficient alternative to their full-sized versions while retaining high accuracy, making them
suitable for practical applications like conversational agents and mental health monitoring tools(Sajid, 2018). The
increasing emphasis on explainability and transparency in AI systems has further highlighted the importance of
visualising emotional sequences, not merely reporting a single sentiment label. This study builds upon this emerging
line of research by applying distilled transformer models to detect sentence-level emotions, quantify emotional
volatility, and visualise emotion drift in a user-friendly interface.

\bigskip
\bigskip

\subsubsection{\textbf{\textcolor{black}{3. Methodology}}}

This study follows a multi-stage methodology designed to detect sentence-level emotions, measure emotional drift within
a text, and evaluate the performance of multiple pre-trained transformer models. The methodology consists of four major
components: dataset preparation, model selection, emotion drift computation and application integration.

\bigskip

{\color[HTML]{666666}
\textbf{\textcolor{black}{Dataset Preparation}}}

\bigskip

The performance of different transformer-based emotion classification models were benchmarked using the publicly
available Emotion Dataset provided by Hugging Face Datasets library. This collection comprises 20,000 text samples with
one of six basic emotions: joy, anger, sadness, fear, love and\enskip{}surprise (Datasets at Hugging Face, 2023). I
choose the dataset because of its balanced quantity, clear label structure, and popularity in emotion recognition.

\bigskip

The dataset was split into training, validation, and test sets as provided. For evaluation purposes, only the test set
(2,000 samples) was used, ensuring that performance metrics reflect the models$\text{\textgreek{’}}$ generalisation to
unseen data. All text samples were preprocessed by lowercasing and removing extraneous whitespace, while preserving
semantic content to maintain the integrity of emotion cues.

\bigskip

{\color[HTML]{666666}
\textbf{\textcolor{black}{Model Selection}}}

\bigskip

Three widely used transformer models were selected to evaluate their effectiveness in sentence-level emotion
classification:

\begin{itemize}[series=listWWNumiv,label=${\bullet}$]
\item DistilRoBERTa\newline

\item DistilBERT\newline

\item DeBERTa\newline

\end{itemize}
The models selected for evaluation were chosen based on their demonstrated effectiveness in emotion classification and
their suitability for sentence-level analysis. DistilRoBERTa was included for its balance of accuracy and efficiency,
making it well-suited for interactive applications. DistilBERT was selected for its consistent and interpretable
predictions, which are particularly valuable for real-time emotion drift analysis. DeBERTa Base represents a more
recent architecture with advanced attention mechanisms capable of capturing subtle emotional nuances, included to
assess whether it could improve accuracy over the distilled models. Together, these models provide a comprehensive
comparison across efficiency, consistency, and state-of-the-art performance, enabling an informed selection for the
application.

\bigskip

The Hugging Face text-classification pipeline was used to perform emotion prediction, enabling consistent inference
across all models.

\bigskip

Each model was evaluated using the same test set and identical processing pipeline to ensure fairness. For every sample
in the test set, the predicted emotion label was compared with the ground truth label. 

\bigskip

Accuracy was used as the comparison metric due to its interpretability and relevance for choosing a model for
deployment.

\bigskip

The evaluation results showed that DistilBERT achieved the highest accuracy (92.7\%), outperforming DistilRoBERTa
(83.9\%) and GoEmotions RoBERTa (19.6\%). Based on this high accuracy and excellent computational efficiency,
DistilBERT was chosen as the foundational model for the final emotion drift analysis application.
\bigskip

{\color[HTML]{666666}
\textbf{\textcolor{black}{Emotion Drift Computation}}}

After selecting the best-performing model, a custom emotion drift pipeline was developed:

\begin{enumerate}[series=listWWNumii,label=\arabic*.,ref=\arabic*]
\item Sentence Segmentation\newline
User input text is split into individual sentences using regex-based segmentation. 
\item Emotion Classification\newline
 Each sentence is passed through the DistilBERT classifier to obtain the predicted emotion label.
\item Emotion Timeline Construction\newline
 The sequence of predicted emotions is arranged chronologically to form an emotion timeline, allowing users to visually
track emotional transitions.

\lfbox[margin=0mm,border-style=none,padding=0mm,vertical-align=top]{\includegraphics[width=4.9953in,height=1.7598in]{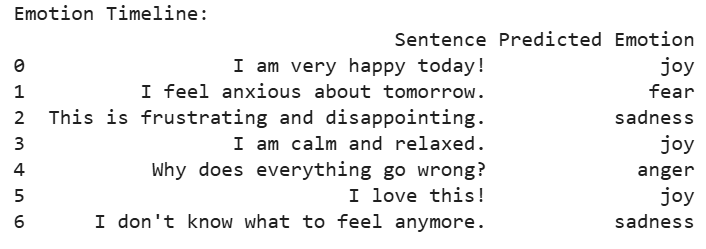}}

\end{enumerate}
{\centering
Figure 1: Emotion timeline of sample text
\par}

\begin{enumerate}[resume*=listWWNumii]
\item Drift Score Calculation\newline

 Emotion drift is computed as the number of emotion changes divided by the total number of transitions.A score of 0
indicates no change in emotional state, while a score close to 1 represents high emotional volatility.\newline

Drift Score = 
$\frac{\mathit{Number}\mathit{of}\mathit{Emotion}\mathit{Changes}}{\mathit{Number}\mathit{of}\mathit{Sentences}-1}$

\bigskip

\item Overall Sentiment Estimation\newline

The final sentiment of the passage is obtained using the DistilBERT model fine-tuned for sentiment analysis(Hugging
Face, 2024) to understand general tone.
\end{enumerate}
\centering
\lfbox[margin=0mm,border-style=none,padding=0mm,vertical-align=top]{\includegraphics[width=2.8693in,height=0.4689in]{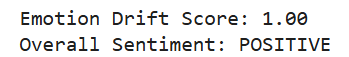}}
\par
{\centering
Figure 2: Emotion Drift Score and Overall Sentiment of text
\par}

\bigskip

\bigskip

\raggedright

{\color[HTML]{666666}
\textbf{\textcolor{black}{Application Integration}}}

\bigskip

The above pipeline was integrated into a Streamlit web application. Users can input any text, view the predicted emotion
timeline, observe the drift score, and receive the overall sentiment. This makes emotion drift analysis accessible and
interpretable to users.

\newpage

{\color[HTML]{434343}

\subsection{\textbf{\textcolor{black}{4. Model Evaluation}}}}
\bigskip

I evaluated three pre-trained transformer models for sentence-level emotion detection.To assess the suitability of
transformer-based architectures for emotion classification, I benchmarked three pre-trained models sourced from the
Hugging Face Model Hub. This phase focused on understanding how different encoder variants perform when applied to
sentence-level emotion detection tasks. 

\bigskip

\lfbox[margin=0mm,border-style=none,padding=0mm,vertical-align=top]{\includegraphics[width=6.2866in,height=4.7339in]{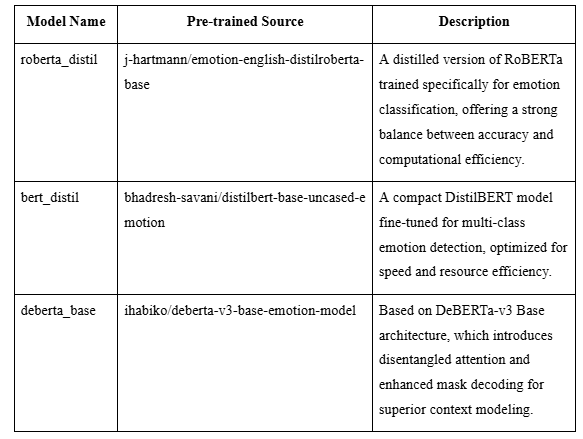}}

{\centering
Figure 3: overview of models selected
\par}
\bigskip

Each model was loaded using the Hugging Face transformers pipeline for text classification, predicting multiple emotions
per sentence. The predicted labels were then converted to a simplified representation for drift analysis.

\newpage

{\color[HTML]{666666}
\textbf{\textcolor{black}{Distil RoBERTa}}}

\bigskip

This is a distilled version of RoBERTa, which means it is a smaller, faster version of the original RoBERTa model that
has been optimized to maintain the majority of its performance.Compared to RoBERTa-base, which has 125M parameters, the
model contains 82M parameters with 6 layers, 768 dimensions, and 12 heads(Hugging Face, 2023).RoBERTa itself is based
on the transformer architecture, using self-attention to create contextualized embeddings for each token in a
sentence(Gandhi, 2025).

\ 
\lfbox[margin=0mm,border-style=none,padding=0mm,vertical-align=top]{\includegraphics[width=7.2902in,height=3.0839in]{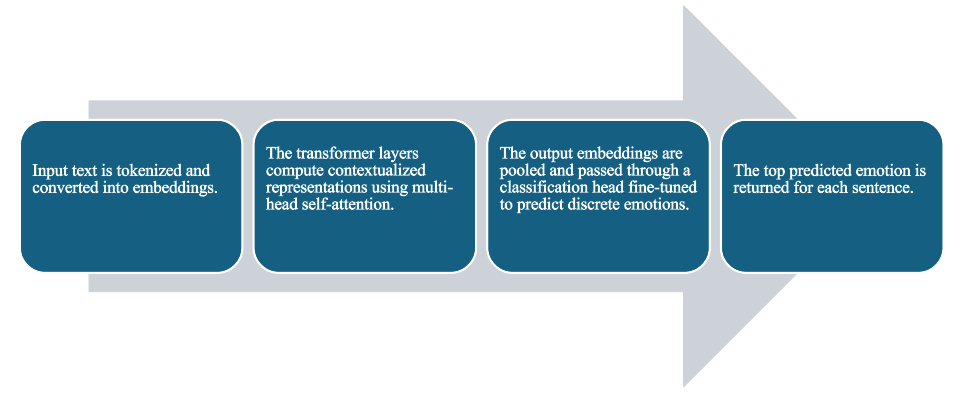}}

{\centering
Figure 4: Workflow for Distil RoBERTa model\newline

\par}

\bigskip

The distilled form retains a high level of general-purpose understanding of language, but it may be less sensitive to
subtle emotional cues due to its diminished depth and representational ability. Because of this, it excels at main or
high-frequency emotions (such as joy, sadness, and anger), but it may have trouble with more complex or ambiguous
emotions like anticipation or confusion.However, this model is a great choice for real-time emotion detection
pipelines.

\newpage

{\color[HTML]{666666}
\textbf{\textcolor{black}{Distil BERT}}}

\bigskip

This model(distilbert-base-uncased-emotion) is a DistilBERT variant, a lighter version of BERT that reduces parameters
and speeds up inference. The “uncased” version ignores capitalization, making it robust to case variations. It is
fine-tuned for emotion detection, providing reliable sentence-level predictions(GeeksforGeeks, 2025).

\bigskip

\lfbox[margin=0mm,border-style=none,padding=0mm,vertical-align=top]{\includegraphics[width=6.5in,height=3.5693in]{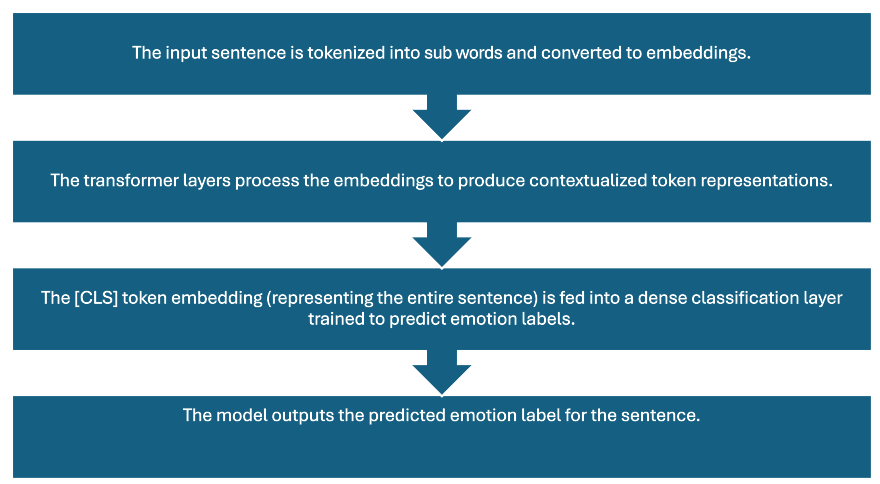}}

{\centering
Figure 5: Workflow for Distil BERT model
\par}

\bigskip

DistilBERT maintains the fundamental advantages of the BERT architecture, such as robust semantic representation and
bidirectional contextual awareness, despite being lightweight(GeeksforGeeks, 2025). This allows it to detect subtle
emotional cues in short text segments, making its predictions both stable and reliable across consecutive sentences. In
the context of emotion drift analysis, this stability ensures that the model captures genuine emotional transitions
rather than producing noisy or inconsistent classifications.

\newpage

{\color[HTML]{666666}
\textbf{\textcolor{black}{DeBERTa}}}

\bigskip

This model uses the DeBERTa-v3 Base architecture, which improves on BERT or RoBERTa by employing disentangled attention
and using enhanced mask decoders (Hugging Face, n.d.). This allows it to capture subtle nuances in text and produce
richer contextual embeddings. The model is fine-tuned for emotion classification, making it suitable for primary
emotion detection.

\bigskip

\lfbox[margin=0mm,border-style=none,padding=0mm,vertical-align=top]{\includegraphics[width=5.8602in,height=4.8744in]{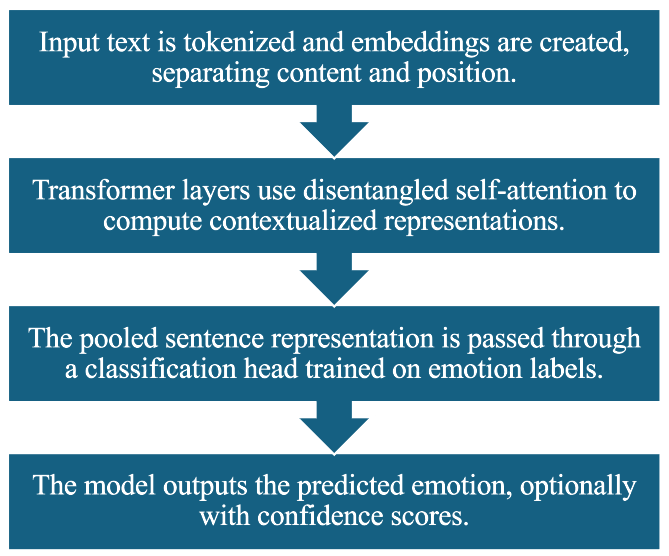}}

{\centering
Figure 5: Workflow for DeBERTa Base model\newline

\par}

While DeBERTa often achieves slightly higher accuracy than distilled models, its predictions can occasionally be
inconsistent for ambiguous sentences due to differences in fine-tuning datasets and label distribution.

\newpage

{\color[HTML]{434343}
\textbf{\textcolor{black}{5. Experiments \& Results}}}

\bigskip

To evaluate the performance of different pre-trained transformer models for sentence-level emotion detection, I choose a
series of text samples\enskip{}and compared their predictions between the three models on sentence-level emotion
detection. The following figure shows the predicted emotions for each sentence across the models:

\bigskip

\lfbox[margin=0mm,border-style=none,padding=0mm,vertical-align=top]{\includegraphics[width=6.5in,height=2.5693in]{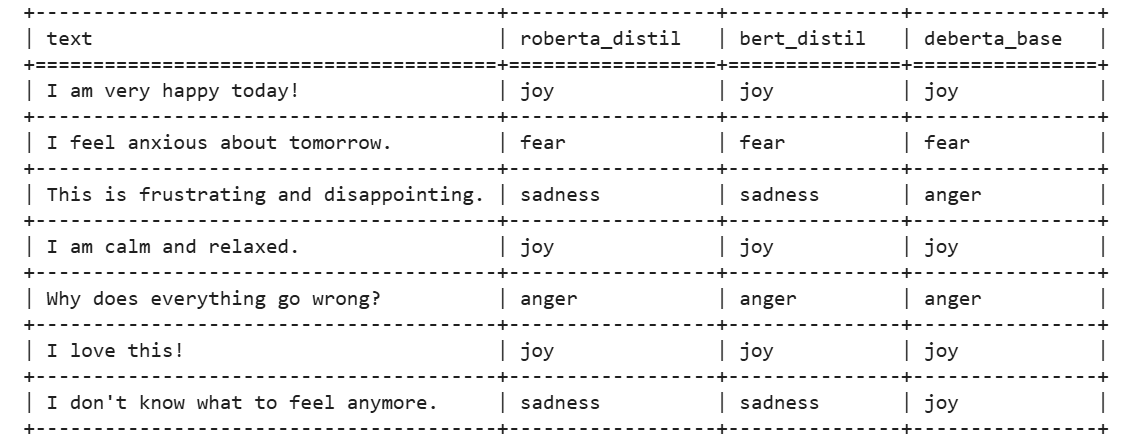}}

{\centering
Figure 6: predicted emotions for sample sentences across models
\par}

\bigskip

The performance of each model was\enskip{}subsequently validated on a bigger benchmark dataset, and the following
results were obtained:

\bigskip

\centering
\lfbox[margin=0mm,border-style=none,padding=0mm,vertical-align=top]{\includegraphics[width=3.0626in,height=1.25in]{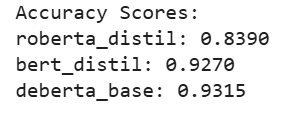}}
\par
{\centering
Figure 7: Accuracy scores of models
\par}

\begin{itemize}[series=listWWNumiii,label=${\bullet}$]
\item roberta\_distil: 0.8390
\item bert\_distil: 0.9270
\item deberta\_base: 0.9315
\end{itemize}

\raggedright

\bigskip

From the evaluation results, deberta\_base achieved the highest overall accuracy (0.9315), slightly outperforming
bert\_distil (0.9270), while roberta\_distil scored 0.8390. Despite a strong accuracy, DeBERTa does\enskip{}indeed
misclassify test sentences sometimes in ways that appear strange or counter-intuitive. For example, categorizing “This
is annoying and disappointing” as anger (rather than sadness) or “I don$\text{\textgreek{’}}$t\enskip{}know what to
feel anymore” as joy.These inconsistencies are likely due to differences in fine-tuning datasets, label schemes, and
sensitivity to subtle emotional cues.

In contrast, bert\_distil produces slightly lower overall accuracy but demonstrates more consistent and intuitive
sentence-level predictions, which is particularly important for real-time emotion drift analysis in interactive
applications. That is why I chose Distil BERT for the Streamlit application as its predictions are more trustworthy to
calculate sentence-level emotional differences.

\bigskip

{\color[HTML]{666666}
\textbf{\textcolor{black}{Example of Model Predictions and Drift Score}}}

\bigskip

In this experiment, I analyzed a short text passage to demonstrate emotion drift detection using three pre-trained
transformer models: roberta\_distil, bert\_distil, and deberta\_base. 

This was the example text used: “I feel overwhelmed today. I tried to reach out for help. Nobody is responding, and I am
frustrated.”

\bigskip

The input text was first split into individual sentences, and each sentence was analyzed to predict its dominant
emotion. The emotion drift score was then computed by calculating the proportion of consecutive sentences that
exhibited a change in emotion. For the example passage, the roberta\_distil model predicted the emotions ['surprise',
'sadness', 'anger'] with a drift score of 1.0, indicating a complete change in emotion across sentences. The
bert\_distil model predicted ['fear', 'fear', 'anger'] with a drift score of 0.5, showing that the first two sentences
maintained the same emotion before changing. The deberta\_base model predicted ['fear', 'joy', 'anger'] with a drift
score of 1.0, again indicating high emotional volatility. This comparison demonstrates how different models capture
emotional transitions, highlighting both the strengths and limitations of each model in detecting nuanced emotional
changes. The drift score effectively quantifies the sequence of emotional changes.

\bigskip

\lfbox[margin=0mm,border-style=none,padding=0mm,vertical-align=top]{\includegraphics[width=6.5in,height=1.6945in]{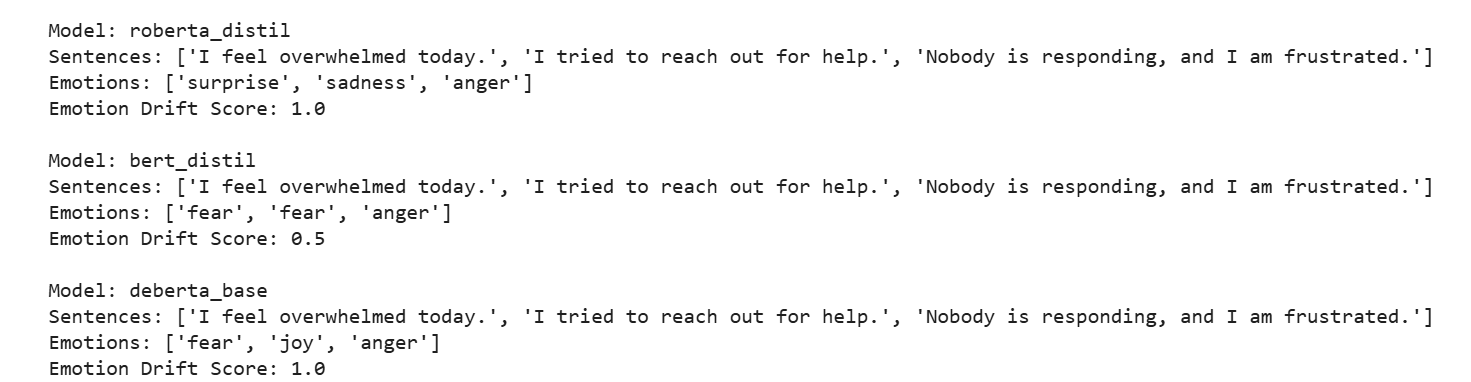}}

{\centering
Figure 8: emotional drift detection across models for a sample text
\par}

\bigskip

The figure below illustrates the emotion drift across three sentences from an example post, as detected by three
different pre-trained models: RoBERTa Distil, BERT Distil, and DeBERTa Base. Each model$\text{\textgreek{’}}$s detected
emotion for every sentence is mapped on the y-axis, while the x-axis represents the sentence sequence (S1–S3). The
lines show how emotions change throughout the text, providing a clear visual representation of emotional transitions.
For instance, RoBERTa Distil identifies a progression from surprise → sadness → anger, indicating a high emotional
drift. BERT Distil shows fear → fear → anger, reflecting a more gradual escalation. DeBERTA Base detects fear → joy →
anger. This visualization highlights how different models interpret emotional changes in text.

\centering
\lfbox[margin=0mm,border-style=none,padding=0mm,vertical-align=top]{\includegraphics[width=5.828in,height=3.0252in]{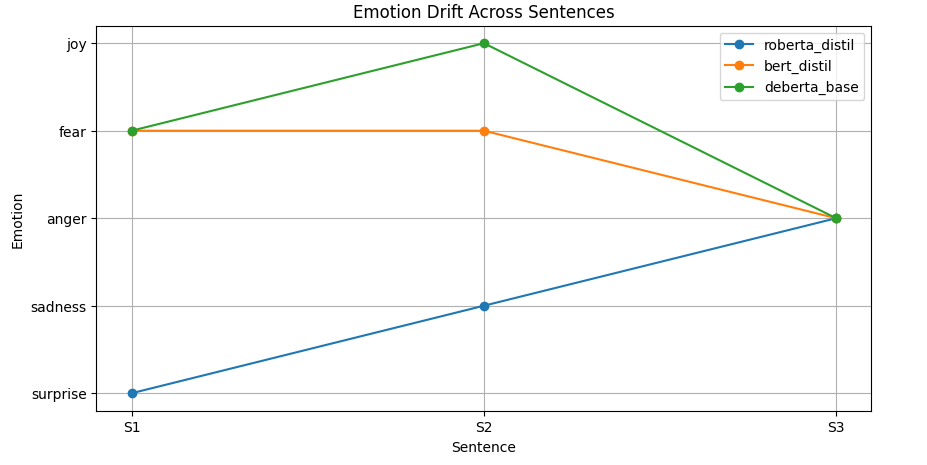}}
\par
{\centering
Figure 9: graph of emotion drift across sentences
\par}

\raggedright

{\color[HTML]{434343}

\subsubsection{\textbf{\textcolor{black}{6. App Development}}}}

\bigskip

A Streamlit application was developed to provide an interactive environment for analyzing emotion drift within
user-submitted text. The app can be accessed here:\url{https://emotion-drift-app.streamlit.app/}.

The system takes a text passage as\enskip{}input, and it divides the text into sentences automatically and uses
pre-trained emotion classifier DistilBERT to obtain sentence-level emotions. Based on performance evaluation,
DistilBERT was considered\enskip{}as the main model for deployment which reported a good accuracy (92.7\%) and
presented reliable and efficient response time for real-time analysis.

The application evaluates how emotions change throughout a piece of text. The input text is broken into smaller
segments, and each segment is classified to generate an Emotion Timeline, showing the sequence of feelings expressed by
the writer. 
\bigskip

\centering
\lfbox[margin=0mm,border-style=none,padding=0mm,vertical-align=top]{\includegraphics[width=2.5161in,height=0.7583in]{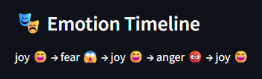}}
\par
\bigskip
{\centering
Figure 10: Emotion Timeline generated in web application
\par}

\bigskip

\raggedright

The degree of emotional fluctuation is then measured using a Drift Score, where higher scores suggest more emotional
volatility.

\centering
\lfbox[margin=0mm,border-style=none,padding=0mm,vertical-align=top]{\includegraphics[width=2.198in,height=0.9063in]{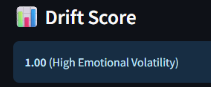}}
\par
\bigskip
{\centering
Figure 11: Drift score generated in web application
\par}

\bigskip

\raggedright

Alongside this, the app also provides an overall sentiment (positive, negative, or neutral) to summarise the dominant
tone of the entire passage. This allows users to visualise both the emotional flow and the general mood of the text in
a clear and intuitive way.\newline

\centering
\lfbox[margin=0mm,border-style=none,padding=0mm,vertical-align=top]{\includegraphics[width=6.5in,height=6.5555in]{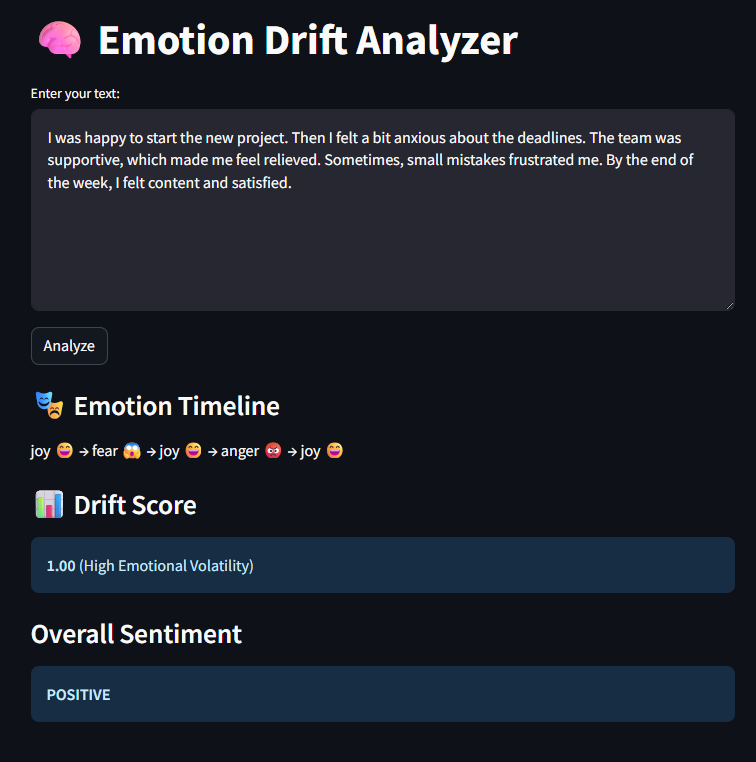}}
\par
\bigskip
{\centering
Figure 12: screenshot of web application
\par}

\bigskip

\raggedright
\newpage

{\color[HTML]{434343}
\subsection{\textbf{\textcolor{black}{7. Conclusion \& Future Work}}}}

The study shows that DeBERTa Base and DistilBERT achieve high accuracy in primary emotion detection, with DeBERTa
slightly outperforming DistilBERT in terms of accuracy.However, DistilBERT offers more consistent and easily
interpretable sentence-level predictions and is a good candidate for\enskip{}real-time usage. Our models both
illustrate how emotion drift scores can capture variation in emotion within a text, while providing more detailed
analysis than traditional sentiment analysis.This approach is applicable to mental health texts, online forums, or
social media, providing a quantitative way to assess emotional dynamics across sequences of sentences.

\bigskip

However the limitation is that the models are trained on general-purpose datasets and may not fully capture
domain-specific emotional expressions in mental health posts.

In conclusion, this paper discusses how to apply transformer models for sentence-level emotion detection to identify
emotion drift in textual data. By computing drift scores, it is easier to quantify emotional variability and highlight
patterns of escalation or relief within posts.

\bigskip

Future work includes:

\bigskip

\begin{itemize}[series=listWWNumi,label=${\bullet}$]
\item Fine-tuning transformer models on domain-specific datasets (e.g., mental health forums) to improve
accuracy.\newline

\item Extending emotion drift analysis to multi-modal data (text + audio/video).\newline

\item Incorporating temporal analysis to study emotion evolution over multiple posts or interactions.
\end{itemize}

\bigskip

\clearpage{\color[HTML]{434343}
\textbf{\textcolor{black}{8. References}}}

\begin{enumerate}[series=listWWNumv,label=\arabic*.,ref=\arabic*]
\item Acheampong, F. A., Nunoo-Mensah, H., \& Chen, Wenyu. (2020, November 28).
Comparative Analyses of BERT, RoBERTa, DistilBERT, and XLNet for Text-based Emotion Recognition.
\nolinkurl{https://www.researchgate.net/publication/346443459_Comparative_Analyses_of_BERT_RoBERTa_DistilBERT_and_XLNet_for_Text-based_Emotion_Recognition}

\item ‌Dilmegani, C. (2024, April 4). Sentiment Analysis Methods Overview, Pros \& Cons. Research.aimultiple.com.
https://research.aimultiple.com/sentiment-analysis-methods/
\item ‌Mitchell, J. (2021). Affective shifts: mood, emotion and well-being. Synthese, 199, 11793–11820.
https://doi.org/10.1007/s11229-021-03312-3
\item ‌Plaza-del-Arco, F. M., Curry, A., Curry, A. C., \& Hovy, D. (2024). Emotion Analysis in NLP: Trends, Gaps and
Roadmap for Future Directions. ArXiv.org. https://arxiv.org/abs/2403.01222
\item ‌Pang, B. and Lee, L. (2008) Opinion Mining and Sentiment Analysis. Foundations and Trends® in Information
Retrieval, 2, 1-135. - References - Scientific Research Publishing. (n.d.). Www.scirp.org.
https://www.scirp.org/reference/referencespapers?referenceid=2442500
\item ‌Liu, M. (2024, June 13). Are mixed emotions real? New research says yes. News and Events.
https://dornsife.usc.edu/news/stories/mixed-emotions-are-real/
\item ‌GoEmotions: A Dataset for Fine-Grained Emotion Classification. (n.d.). Research.google.
https://research.google/blog/goemotions-a-dataset-for-fine-grained-emotion-classification/
\item ‌Christ, L., Amiriparian, S., Milling, M., Aslan, I., \& Schuller, B. W. (2024). Modeling Emotional Trajectories
in Written Stories Utilizing Transformers and Weakly-Supervised Learning. ArXiv.org. https://arxiv.org/abs/2406.02251
\item ‌Sajid, H. (2018). Distilbert: A Smaller, Faster, and Distilled BERT - Zilliz blog. Zilliz.com.
https://zilliz.com/learn/distilbert-distilled-version-of-bert
\item ‌dair-ai/emotion $\cdot $ Datasets at Hugging Face. (2023, March 23). Huggingface.co.
https://huggingface.co/datasets/dair-ai/emotion
\item ‌j-hartmann/emotion-english-distilroberta-base $\cdot $ Hugging Face. (n.d.). Huggingface.co.
https://huggingface.co/j-hartmann/emotion-english-distilroberta-base
\item ‌bhadresh-savani/distilbert-base-uncased-emotion $\cdot $ Hugging Face. (n.d.). Huggingface.co.
https://huggingface.co/bhadresh-savani/distilbert-base-uncased-emotion
\item ihabiko/deberta-v3-base-emotion-model $\cdot $ Hugging Face. (2025). Huggingface.co.
https://huggingface.co/ihabiko/deberta-v3-base-emotion-model
\item ‌distilbert/distilbert-base-uncased-finetuned-sst-2-english $\cdot $ Hugging Face. (2024, January 30).
Huggingface.co. https://huggingface.co/distilbert/distilbert-base-uncased-finetuned-sst-2-english
\item distilbert/distilroberta-base $\cdot $ Hugging Face. (2023, April 5). Huggingface.co.
https://huggingface.co/distilbert/distilroberta-base
\item ‌Gandhi, D. (2025, May 26). RoBERTa Model Explained: Features, Benefits \& Use Cases. Dhiwise.com.
https://www.dhiwise.com/post/roberta-model
\item ‌GeeksforGeeks. (2025, June 26). Introduction to DistilBERT Model. GeeksforGeeks.
https://www.geeksforgeeks.org/nlp/introduction-to-distilbert-model/
\item ‌microsoft/deberta-v3-base $\cdot $ Hugging Face. (n.d.). Huggingface.co.
https://huggingface.co/microsoft/deberta-v3-base
\end{enumerate}

\bigskip

\bigskip
\end{document}